\documentclass{article}

\usepackage{microtype}
\usepackage{graphicx}
\usepackage{subfigure}
\usepackage{booktabs}
\usepackage{hyperref}

\usepackage[accepted]{icml2022}

\usepackage{amsmath}
\usepackage{amssymb}
\usepackage{mathtools}
\usepackage{amsthm}

\usepackage[capitalize,noabbrev]{cleveref}

\usepackage{physics}
\usepackage{multirow}
\usepackage{url}

\theoremstyle{plain}
\newtheorem{theorem}{Theorem}[section]

\theoremstyle{definition}
\newtheorem{definition}[theorem]{Definition}

\theoremstyle{remark}

\newcommand{\eg}[0]{\textit{e.g.}}
\newcommand{\etal}[0]{\textit{et~al.}}
\newcommand{\ie}[0]{\textit{i.e.}}
\newcommand{\ul}[1]{\underline{#1}}
\newcommand{\tbf}[1]{\textbf{#1}}
\newcommand{\ulbf}[1]{\underline{\tbf{#1}}}

\newcommand{\fig}[3]{
  \begin{figure}[t]
    \begin{center}
      \centerline{\includegraphics[width=\columnwidth]{fig/#1.pdf}}
      \caption{#2}
      \label{#3}
    \end{center}
    \vskip -0.2in
  \end{figure}
}

\newcommand{\tab}[5][]{
  \begin{table#1}[t]
    \caption{#2}
    \label{#3}
    \begin{center}
    \begin{small}
    \begin{sc}
      \begin{tabular}{#4}
        #5
      \end{tabular}
    \end{sc}
    \end{small}
    \end{center}
    \vskip -0.1in
  \end{table#1}
}

\newcommand{\T}[1]{\text{#1}}
\newcommand{\argmax}{\mathop\T{argmax}\nolimits}

\newcommand{\LSE}{\mathop\T{LSE}\nolimits}
\newcommand{\N}[0]{\mathbb{N}}
\newcommand{\R}[0]{\mathbb{R}}

\newcommand{\M}[0]{\mathcal{M}}
\renewcommand{\S}[0]{\mathcal{S}}
\newcommand{\X}[0]{\mathcal{X}}
\newcommand{\Y}[0]{\mathcal{Y}}
\newcommand{\e}[1]{{\exp\qty(#1)}}

\newcommand{\x}[0]{\vb{x}}
\newcommand{\xa}[0]{{\x^\ast}}

\newcommand{\nablax}[0]{{\vb*{\nabla}_{\x}}}
\newcommand{\xy}[0]{{\qty(\x,y)}}
\newcommand{\ellT}[1]{{\ell_{\T{#1}}}}
\newcommand{\ellTxy}[1]{{\ellT{#1}\xy}}
\newcommand{\ellxy}[0]{{\ell\xy}}
\newcommand{\Yby}[0]{{\Y\backslash\qty{y}}}
\newcommand{\SY}[0]{{\S\qty(\Y)}}
\newcommand{\Sy}[0]{{\S\qty(y)}}
\newcommand{\Smy}[0]{{\S^-\qty(y)}}

\newcommand{\pii}[1]{{\pi\qty(#1)}}
\newcommand{\Mstd}[0]{{\M_{\T{std}}}}
\newcommand{\Madv}[0]{{\M_{\T{adv}}}}

\icmltitlerunning{Superclass Adversarial Attack}

\begin{document}

\twocolumn[
\icmltitle{Superclass Adversarial Attack}

\begin{icmlauthorlist}
\icmlauthor{Soichiro Kumano}{ut}
\icmlauthor{Hiroshi Kera}{cu}
\icmlauthor{Toshihiko Yamasaki}{ut}
\end{icmlauthorlist}

\icmlaffiliation{ut}{Graduate School of Information Science and Technology, The University of Tokyo, Hongou 7-3-1, Bunkyoku, 113-8656, Tokyo, Japan.}
\icmlaffiliation{cu}{Graduate School of Engineering, Chiba University, Yayoityo 1-33, Chiba, 263-8522, Chiba, Japan.}

\icmlcorrespondingauthor{Soichiro Kumano}{kumano@cvm.t.u-tokyo.ac.jp}

\icmlkeywords{Adversarial Attack, Adversarial Examples, Superclass}

\vskip 0.3in
]

\begin{abstract}
  Adversarial attacks have only focused on changing the predictions of the classifier, but their danger greatly depends on how the class is mistaken. For example, when an autonomous driving system mistakes a Persian cat for a Siamese cat, it is hardly a problem. However, if it mistakes a cat for a 120km/h minimum speed sign, serious problems can arise. As a stepping stone to more threatening adversarial attacks, we consider the superclass adversarial attack, which causes misclassification of not only fine classes, but also superclasses. We conducted the first comprehensive analysis of superclass adversarial attacks~(an existing and 19 new methods) in terms of accuracy, speed, and stability, and identified several strategies to achieve better performance. Although this study is aimed at superclass misclassification, the findings can be applied to other problem settings involving multiple classes, such as top-k and multi-label classification attacks.
\end{abstract}

\section{Introduction}
Adversarial attacks aim to fool a classifier to misclassify~\cite{AEFirst, FGSM,InversionAttack,DeepFool,JSMA,BIM,CWAttack,UAP,EAD,PGD,WithMomentum,BreakICLR18Attack,OnePixelAttack,AutoAttack,FAB,SquareAttack}. These attacks appear not only in the electronic world, but also in the physical world, and create a serious risk for systems with deep neural networks~(DNNs)~\cite{AEGlasses,AEPhysicalFirst,StopSign,turtle,AEGlasses2,HumanWithAE,AdvFashion}. However, prior studies have not evaluated which classes should be targeted for more harmful attacks, and thus not all the classification errors by the attacks are risky for DNN-based systems. For example, if an attack on a DNN-based autonomous driving system causes that system to mistake a Persian cat for a Siamese cat, the error is trivial. However, if the cat is instead mistaken for a 120km/h minimum speed sign, serious accidents may occur. Therefore, attacks that cause erroneous classification may have widely varying risk levels.

\fig{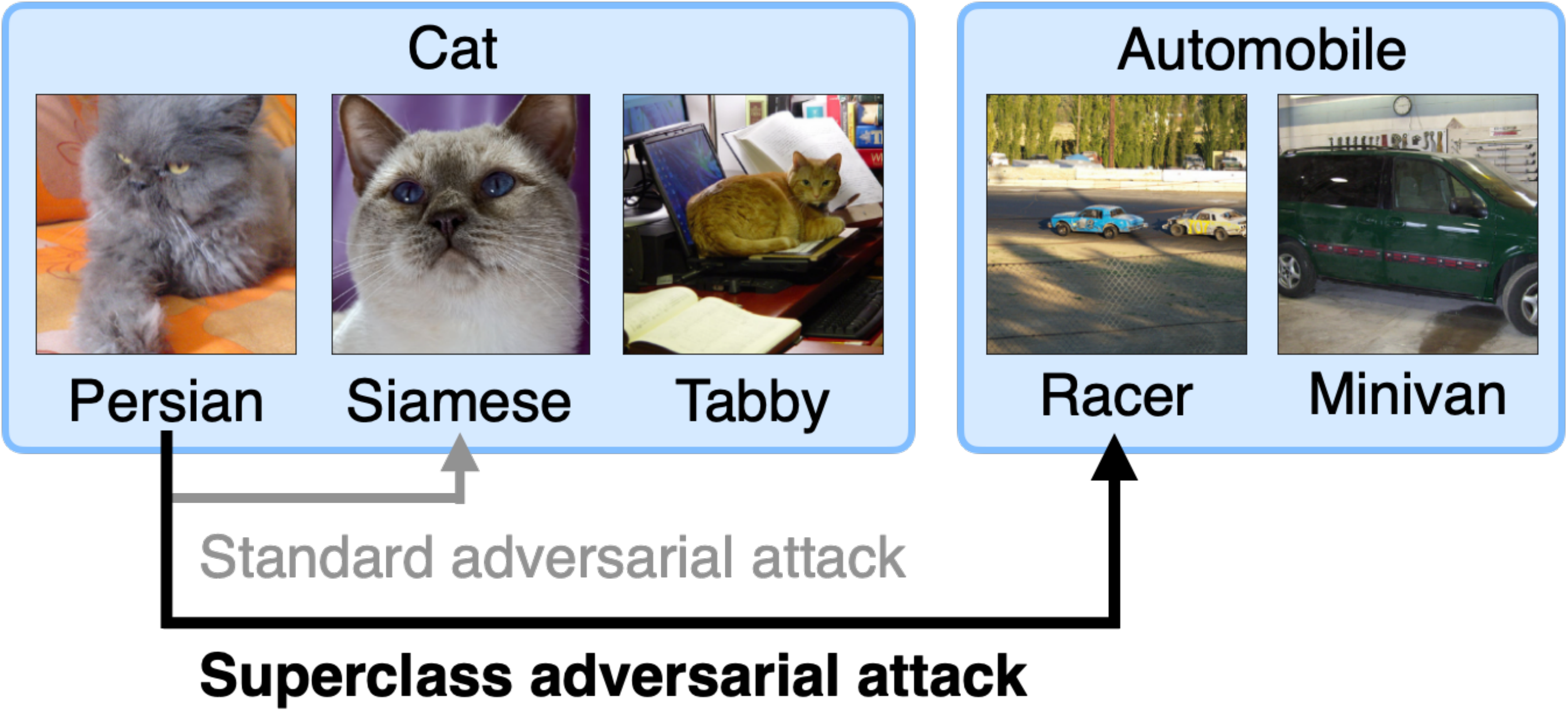}{Superclass adversarial attacks result in misclassification of not only fine classes, but also superclasses. Adversarial examples generated by a superclass attack can be misclassified into semantically distant classes, which may cause serious problems.}{fig:abst}

In this study, to explore adversarial attacks on a more threatening level, we consider the superclass adversarial attack~(SAA). Superclass adversarial examples~(SAEs) generated by SAAs are misclassified to different superclasses. When superclasses are defined by semantic similarities, SAEs are more semantically distant from the original class than standard adversarial examples. For example, given a sample of Persian cat, its SAE will be misclassified as a sedan or minivan, not as a nearby class such as a Siamese or tabby cat~(\cref{fig:abst})\footnote{Note that this study is valid for any superclass configuration. There are other settings besides visual similarities, such as dangerousness.}. Thus, SAEs cause more serious misclassification for DNNs than standard adversarial attacks. We conducted a first comprehensive analysis of SAAs, including suited attack frameworks, efficient loss functions, and the inner mechanisms of SAAs, which have not been extensively researched in prior studies.

We first show that standard adversarial attacks cause misclassification between similar classes, typically within the same superclasses~(\eg, from a Persian cat to a Siamese cat), and thus are often not effective threats in real-world systems. Next, we reveal that the existing iterative sequence-based SAA~\cite{WorstTargetedAttack} is not practical in terms of time efficiency, and we propose an alternative SAA. The iterative sequence-based SAA repeats a targeted attack to each fine class until success, which is highly time-intensive. In contrast, our proposed iterative sort-based SAA leverages logit-based sort and exhibits faster performance without loss of accuracy. As a practical compromise, we also incorporate early-stopping into the iterative sort-based SAA. Instead of repeating targeted attacks, we propose an efficient non-targeted attack that aims to increase the loss for classes within the same superclass. We investigate 15 different types of such losses, including the simple sum of cross-entropy~(CE) loss. Because the success rate of gradient-based adversarial attacks~(\eg, projected gradient descent~(PGD)~\cite{PGD}) strongly depends on the loss function~\cite{CWAttack,AutoAttack}, it is important to verify multiple loss functions. The proposed 15 loss functions are categorized into max-, sum-, and LogSumExp~(LSE)-based methods. The max-based method decreases the maximum probability in the original superclass at each step, which gradually creates a decrease in the superclass’s overall probability. As methods that can be run in a single step~(\eg, fast gradient sign method~(FGSM)~\cite{FGSM}), we propose sum- and LSE-based methods, which are constructed by the sum and LSE of fine class logits, probability, or CE, respectively. These methods calculate the gradients of multiple classes, and decrease multiple probabilities in an original superclass in a single step. In the results of comprehensive experiments for these non-iterative methods, we have identified that certain variants of LSE- and sum-based methods yield stronger attacks in single- and multi-step, respectively. We also found that loss based on CE in non-iterative SAAs does not work well due to conflicts between multiple CE and an accelerated gradient vanishing problem. These findings are not limited to SAAs, and are applicable to general attacks involving multiple classes, including top-k and multi-label classification attacks. Our contributions are summarized as follows.
\begin{itemize}
  \item We conducted a comprehensive analysis of SAAs, which have received little focus in literature.
  \item We propose an iterative sort-based SAA, which is significantly faster than the existing method, while being comparably strong. In addition, we propose the implementation of early-stopping in the sort-based method as a realistic compromise.
  \item We propose non-iterative SAAs with 15 loss functions based on maximum, sum, and LSE strategies. Comprehensive evaluations show that a variant of the LSE-based method in a single-step, and that of the sum-based method in multi-steps, yield the highest success rates and efficiency in non-iterative methods.
  \item We discuss three strategies and several limitations with cross-entropy loss in the context of non-iterative SAAs. These results not only facilitate the future development of SAAs, but can also be applied across various problem settings involving multiple classes.
\end{itemize}

\section{Related Work}
\label{sec:related}
FGSM and PGD generate adversarial examples by causing perturbations, which are calculated by gradients of natural images to increase the loss~(\eg, CE loss). Such gradient-based attacks are strongly affected by the design of the loss function~\cite{CWAttack,AutoAttack}. Using this study as a baseline, we propose and compare a variety of loss functions for SAAs. 

The standard non-targeted adversarial attack does not distinguish the class to be misclassified. This can lead to attacks that pose little threat to real systems. In contrast, an SAA causes the misclassification of not only the fine class, but also the superclass. To SAEs, Ma~\etal~performed targeted attacks sequentially on the fine classes in different superclasses~\cite{ma2018characterizing}. This method is very slow due to its structure consisting of repeated targeted attacks. We performed a comprehensive analysis of SAAs, including the existing method, its modifications, and novel methods, in terms of accuracy and time consumption.

Top-k and multi-label classification attacks are also considered as adversarial attacks involving multiple classes~\cite{song2018multi,zhou2020generating,tursynbek2022geometry,zhang2020learning,hu2021tkml,song2018multi,zhou2020generating,ghamizi2021evasion,melacci2021domain,yang2021characterizing,yang2021attack,zhou2021hiding}. In \cref{sec:ApRelated}, these two attacks are discussed in detail.

\section{Methods}
This section describes existing and our proposed SAA methods, as summarized in~\cref{tab:methods}.

\tab[*]{Comparison table of standard non-targeted attacks and SAAs. Our proposed SAAs are indicated in bold. Loss $\ellxy$ is used in \cref{eq:AdvAtk}. Iterative SAAs repeat targeted attacks until success, whereas non-iterative SAAs perform a single non-targeted attack.}{tab:methods}{lllll}{
  \toprule
  \multicolumn{3}{l}{attack type} & loss type & loss $\ellxy$ \\ \midrule
  \multicolumn{3}{l}{standard non-targeted attack} & ce &$-\ln p_y$ \\ 
  & & & cw~\cite{CWAttack} & $-l_y+l_\pii{\Yby}$ \\ 
  & & & prob-cw & $-p_y+p_\pii{\Yby}$ \\ 
  & & & weighted-cw & $-p_yl_y+p_\pii{\Yby}l_\pii{\Yby}$ \\ \midrule
  saa & iterative & sequential & ce~\cite{WorstTargetedAttack} & $\ln p_i~\qty(i\in\Smy)$ \\ \cmidrule(lr){3-5}
  & & \tbf{sorted} & \tbf{ce} & $\ln p_i~\qty(i\in\qty[\Smy]^k)$ \\
  & & & \tbf{cw} & $l_i-l_\pii{\Y\backslash\{i\}}~\qty(i\in\qty[\Smy]^k)$ \\
  & & & \tbf{prob-cw} & $p_i-p_\pii{\Y\backslash\{i\}}~\qty(i\in\qty[\Smy]^k)$ \\
  & & & \tbf{weighted-cw} & $p_il_i-p_\pii{\Y\backslash\{i\}}l_\pii{\Y\backslash\{i\}}~\qty(i\in\qty[\Smy]^k)$ \\ \cmidrule(lr){2-5}
  & \tbf{non-iterative} & \tbf{max} & \tbf{ce} & $-\ln p_\pii{\Sy}$ \\
  & & & \tbf{cw} & $-l_\pii{\Sy}+l_\pii{\Smy}$ \\
  & & & \tbf{prob-cw} & $-p_\pii{\Sy}+p_\pii{\Smy}$ \\
  & & & \tbf{weighted-cw} & $-p_\pii{\Sy}l_\pii{\Sy}+p_\pii{\Smy}l_\pii{\Smy}$ \\ \cmidrule(lr){3-5}
  & & \tbf{sum} & \tbf{ce} & $-\sum_{i\in\Sy}\ln p_i$ \\
  & & & \tbf{logit-ce} & $-\ln \frac{\e{\sum_{i\in\Sy}l_i}}{\sum_{I\in\SY}\e{\sum_{i\in I}l_i}}$ \\
  & & & \tbf{prob-ce} & $-\ln\sum_{i\in\Sy}p_i$ \\
  & & & \tbf{cw} & $-\sum_{i\in\Sy}l_i+l_\pii{\Smy}$ \\
  & & & \tbf{prob-cw} & $-\sum_{i\in\Sy}p_i+p_\pii{\Smy}$ \\
  & & & \tbf{weighted-cw} & $-\sum_{i\in\Sy}p_il_i+p_\pii{\Smy}l_\pii{\Smy}$ \\ \cmidrule(lr){3-5}
  & & \tbf{LSE} & \tbf{ce} & $\LSE_{i\in\Sy}\qty(-\ln p_i)$ \\
  & & & \tbf{prob-ce} & $-\ln\LSE_{i\in\Sy}p_i$ \\
  & & & \tbf{cw} & $-\LSE_{i\in\Sy}l_i+l_\pii{\Smy}$ \\
  & & & \tbf{prob-cw} & $-\LSE_{i\in\Sy}p_i+p_\pii{\Smy}$ \\
  & & & \tbf{weighed-cw} & $-\LSE_{i\in\Sy}p_il_i+p_\pii{\Smy}l_\pii{\Smy}$ \\
  \bottomrule
}

\subsection{Notations and Definitions}
\label{sec:NotaDef}
We denote the set of images by $\X$, and the set of fine classes by $\Y:=\{1,\ldots,K\}$. A map $S:\Y\to\mathcal{P}(\Y)$ accepts a fine class as input, and returns its super class, where $\mathcal{P}(\Y)$ is the power set of $\Y$. Extending our notation, we denote the set of superclasses by $\SY:=\{\S(i)\mid i\in\Y\}$. Note that we assume that any two different superclasses are disjoint~(\ie, for any $S_1,S_2\in\SY$ with $S_1\ne S_2$, it holds that $S_1\cap S_2=\emptyset$). The logit and probability of input $\x$ with respect to the fine class $i\in\Y$ are denoted by $l_i(\x)$ and $p_i(\x)$, respectively. When the input is not of our interest, we write $l_i:=l_i(\x)$ and $p_i:=p_i(\x)$ for simplicity. The prediction of classifier $f:\X\to\Y$ for $\x\in\X$ is $f(\x):=\argmax_{i\in\Y}l_i(\x)$. For simplicity, we define $\Smy:=\Y\backslash\Sy$ and $\pii{A}:=\argmax_{i\in\mathcal{A}}l_i$ for $A\subset\Y$. In addition, we denote by $\qty[A]^k$ the subset of $A\subset\Y$ corresponding to the $k$ largest logits.

While adversarial examples are misclassified between different fine classes, superclass adversarial examples~(SAEs) are misclassified between the fine classes of different superclasses. Formally, an SAE is defined as follows:
\begin{definition}[superclass adversarial example]
\label{def:SAE}
A sample $\xa\in\X$ is called a superclass adversarial example of an image $\x\in\X$~(with label $y\in\Y$) if it satisfies $f(\xa)\notin\Sy$ and $d(\x,\xa)\le\epsilon$, where $d:\X\times\X\to\R_{\geq0}$ and $\epsilon>0$ denote a distance function and constraint, respectively.
\end{definition}
\noindent When $\Sy=\{y\}$, SAEs are reduced to standard adversarial examples. We also refer to the methods of generating SAEs as superclass adversarial attacks~(SAAs).

\subsection{Iterative SAA}
\label{sec:MethodIterative}
An iterative SAA runs the targeted attack on every fine class $i\in\Smy$ that is not included in the original superclass $\Sy$ until success. Ma~\etal~were the first to propose iterative SAAs~\cite{WorstTargetedAttack}. Because they did not describe a selection method for the order of targeted fine classes to attack, we assume that they used the ascending order of the fine class index. We refer to this method as the iterative sequence-based method~(worst-case targeted attack in their study). In this paper, we propose a new iterative SAA called the iterative sort-based method, which selects target fine classes in the descending order of the logit $l_i(\x)~(i\in\Smy)$. Because empirically, the larger the original logit is, the more likely the targeted attack will succeed, the sort-based method performs faster than the sequence-based method in almost all cases. Even in the worst case, the sort-based method attacks all targeted classes, and thus does not lose the attack success rate compared to the sequence-based method.

Although the sort-based method exhibits faster performance than the sequence-based method, it is still slower than a standard attack due to the nature of repeated targeted attacks. Therefore, we propose to incorporate early-stopping into the sort-based method as a practical compromise. Instead of attacking all possible targeted fine classes $i\in\Smy$, the modified method would attack only the $k\in\N$ fine classes $i\in\qty[\Smy]^k$, whose logit $l_i(\x)~(i\in\Smy)$ is the largest. Although larger values of $k$ improve accuracy, they also increase the time cost. In this study, we use CE, CW, prob-CW~(see below), and weighted-CW~(see below) losses as loss functions in PGD.

\tab[*]{Overview table of SAAs. Values represent superclass accuracy~(\%) and runtime~(minutes).}{tab:overview}{llrrrr}{
  \toprule
  \multirow{2}{*}{attack type} & \multirow{2}{*}{method} & \multicolumn{2}{r}{cifar-100} & \multicolumn{2}{r}{imagenet} \\ \cmidrule(lr){3-4} \cmidrule(lr){5-6}
  & & $\Mstd$ & $\Madv$ & $\Mstd$ & $\Madv$ \\ \midrule
  n/a & n/a & 87.5\%~(0.00min) & 76.5\%~(0.00min) & 81.7\%~(0.00min) & 64.2\%~(0.00min) \\
  standard & non-targeted & 26.5\%~(1.00min) & 43.1\%~(2.00min) & 12.2\%~(1.00min) & 26.1\%~(2.00min) \\ \midrule
  saa & iterative & 0.00\%~(0.00min) & 26.1\%~(4.00min) & 0.00\%~(0.00min) & 21.5\%~(5.00min) \\
  & non & 0.00\%~(0.00min) & 26.3\%~(1.00min) & 0.00\%~(0.00min) & 21.7\%~(2.00min) \\
  \bottomrule
}

\subsection{Non-Iterative SAA}
\label{sec:MethodSingle}
The non-iterative method is an extension of the standard non-targeted attack framework for the SAA. This method is considerably faster than the iterative methods, as it does not repeat attacks. The foundation of the non-iterative method is to reduce the probability of all fine classes in the original superclass $\Sy$. In this paper, we introduce three different strategies for the non-iterative method: max, sum, and LSE.

We briefly introduce a common framework for gradient-based standard attacks and SAAs. Gradient-based attacks generate the adversarial example $\xa$ by updating the image $\x\in\X$ in the fine class $y\in\Y$ to increase the loss $\ellxy$ as follows:
\begin{align}
  \label{eq:AdvAtk}
  \xa:=\Pi_{d(\x,\xa)\leq\epsilon}\qty{\x+\alpha\cdot\nablax\ellxy},
\end{align}
where $\Pi_{d(\x,\xa)\leq\epsilon}:\X\to\X$ is a projection operator, $\alpha>0$ is the step size, and $\epsilon>0$ is the distance constraint. \cref{eq:AdvAtk} indicates that the loss function’s design determines the nature of the attack. It is established that attack success rate varies significantly depending on the loss function~\cite{CWAttack,AutoAttack}. In particular, the CE and CW functions~\cite{CWAttack} are common loss functions in adversarial attacks. These functions are expressed as follows:
\begin{align}
  \ellTxy{CE}&:=-\ln p_y.\\
  \ellTxy{CW}&:=-l_y+l_\pii{\Y\backslash\{y\}}.
\end{align}

In addition, we propose the prob-CW and weighted-CW as new loss function baselines, defined as follows:
\begin{align}
  \ellTxy{PCW}&:=-p_y+p_\pii{\Y\backslash\{y\}},\\
  \ellTxy{WCW}&:=-p_yl_y+p_\pii{\Y\backslash\{y\}}l_\pii{\Y\backslash\{y\}}.
\end{align}
We extend CE, CW, prob-CW, and weighted-CW to the loss functions in max-, sum-, and LSE-based methods for SAAs~(\cref{tab:methods}). Henceforth, the combination of the max-based method and CE loss is referred to as max~(CE). Other combinations are similarly named~(\eg, sum~(CW) and LSE~(prob-CW)). Although all of these loss functions appear to have similar roles, we show in \cref{sec:exp} that they produce different results in terms of robust accuracy.

\paragraph{Max-Based Method.}
In each step, the max-based method takes $\pii{\Sy}$, the class with the highest logit among all fine classes in the original superclass $\Sy$, and decreases its probability. This is considered relatively easy to optimize, as one does not need to consider the gradients of multiple classes simultaneously.

\paragraph{Sum-Based Method.}
The sum-based method sums the values~(CE, logits, probability, or products of logit and probability) of all fine classes $i\in\Sy$ in the original superclass $\Sy$. Note that the second terms of extended CW, prob-CW, and weighted-CW loss~(17, 18, and 19-th rows in~\cref{tab:methods}), which are incentives to increase the probability of a fine class $i\in\Smy$ in a different superclass, should not use multiple class values. This is because there is no advantage in SAA to increasing the probability of multiple fine classes. Furthermore, the sum- and LSE-based methods offer several ways to calculate extended CE loss. Sum~(logit-CE) calculates the probability in CE from the logit of the superclass and the softmax function. The logit of superclass is defined as $\sum_{i\in I}l_i$ for superclass $I\in\SY$~(\cref{tab:methods}). Sum~(prob-CE) defines the probability of a superclass as $\sum_{i\in I}p_i$ for superclass $I\in\SY$~(\cref{tab:methods}). 

\paragraph{LSE-Based Method.}
The LSE-based method uses the LogSumExp~(LSE) function. Note that the loss function of LSE~(logit-CE) can be rearranged to that of sum~(prob-CE); thus, these two are equivalent. This function differs from the sum function in that the larger elements are given a higher weighting. For example, for the first term of LSE~(CW), the gradients of logits are weighted as $\sum_{i\in\Sy}\frac{\e{l_i}}{\sum_{i\in\Sy}\e{l_i}}\nablax l_i$; in contrast, they are not weighted in sum~(CW) as $\sum_{i\in\Sy}\nablax l_i$. This ensures strong attacks on classes with large probabilities and gives little consideration to classes with small probabilities. Note that other sum-based methods~(prob-CE, prob-CW, and weighted-CW) weigh the gradients differently.

\section{Experimental Results}
\label{sec:exp}
In this paper, superclass accuracy is defined as the ratio of data for which the predicted superclass matches the ground truth. We used CIFAR-100~\cite{CIFAR10} and 5,000 images extracted from ImageNet~(ILSVRC2012)~\cite{ImageNet}. The superclasses were constructed by visual similarity between the fine classes. Superclass configurations followed~\cite{CIFAR10,ImageNetSuperclass}. The average number of elements in each superclass was 5.00 in CIFAR-100 and 1.79 in ImageNet. As classifiers, two WideResNet~\cite{WideResNet} were used, standard trained $\Mstd$, and adversarial trained $\Madv$ for each dataset. Unless stated otherwise, the PGD settings for evaluation were 100 steps, 8/255 epsilons, 2/255 step size, $l_{\infty}$ norm. All SAEs terminated upon successful attack. Additional configuration details are described in \cref{sec:ApSettings}.

\tab[*]{Comparison table of iterative SAAs with CE loss. Values represent accuracy~(\%) and time~(minutes). $k$ is a hyper-parameter for early-stopping. $k=\abs{\Smy}$ indicates that early-stopping is not used, and all target labels can be attacked~(cf. \cref{sec:MethodIterative}).}{tab:iterative}{llrrrr}{
  \toprule
  \multirow{2}{*}{method} & \multirow{2}{*}{$k$} & \multicolumn{2}{r}{cifar-100} & \multicolumn{2}{r}{imagenet} \\ \cmidrule(lr){3-4} \cmidrule(lr){5-6}
  & & $\Mstd$ & $\Madv$ & $\Mstd$ & $\Madv$ \\ \midrule
  sorted & 1 & 0.00\%~(0.00min) & 28.2\%~(1.00min) & 0.00\%~(0.00min) & 27.5\%~(2.00min) \\
  & 3 & 0.00\%~(0.00min) & 26.1\%~(4.00min) & 0.00\%~(0.00min) & 22.5\%~(6.00min) \\
  & 5 & 0.00\%~(0.00min) & 25.7\%~(7.00min) & 0.00\%~(0.00min) & 21.9\%~(9.00min) \\
  & $\abs{\Smy}$ & 0.00\%~(0.00min) & 25.4\%~(138min) & 0.00\%~(0.00min) & 20.6\%~(153$\times10^1$min) \\ \midrule
  seq.~\cite{WorstTargetedAttack} & $\abs{\Smy}$ & 0.00\%~(0.00min) & 25.4\%~(170min) & 0.00\%~(0.00min) & 20.6\%~(203$\times10^1$min) \\
  \bottomrule
}

\subsection{Overview of SAAs}
\label{sec:overview}
Superclass accuracy and runtime are summarized in \cref{tab:overview}. The superclass accuracy was extracted from the best results of each method. The hyper-parameter $k$ of the iterative SAA is the minimum among those whose robust accuracy is greater than the best accuracy of the non-iterative SAA. Detailed results are in \cref{tab:ApStd,tab:ApIterative,tab:ApNon}

\cref{tab:overview} shows that standard non-targeted attacks cannot cause superclass misclassification when the superclasses are constructed by visual similarities. Compared to CIFAR-100, the difference in superclass accuracy between standard attacks and SAAs is slightly smaller in ImageNet. This is due to the average number of elements in each superclass of ImageNet being smaller than that of CIFAR-100, and even standard attacks tended to cause superclass misclassification. Since the update of adversarial examples was stopped upon successful superclass misclassification, some standard attacks were slower than SAAs.

In contrast to standard non-targeted attacks, SAAs can induce superclass misclassification. Although the iterative method with an appropriate hyper-parameter $k$ can achieve a higher success rate than the non-iterative method, it remains more time-intensive. Therefore, the iterative method is recommended for strong attacks where time is not a concern. Note, however, that we do not recommend that the evaluation for adversarial robustness of classifiers with a large $k$ be the standard in the field~(cf. \cref{sec:ExpIterative}). Because the non-iterative method is fast and achieves high rates of success, it can be employed for real-time attacks and adversarial training that necessitate high speed.

\subsection{Comparison of Iterative Method}
\label{sec:ExpIterative}
\cref{tab:iterative} shows the results of iterative SAAs with CE loss. Note that the values in \cref{tab:iterative} are different from those in \cref{tab:overview}, where the best loss function was selected among several. More detailed results for iterative SAAs are listed in \cref{tab:ApIterative}.

For the standard trained models $\Mstd$, there are no significant differences in robust accuracy between all methods. This is because the standard trained models are very vulnerable to attacks, which succeed regardless of the first targeted class. Instead, major differences between the methods are seen in the adversarial robust models $\Madv$. Because these models are robust to attacks, the runtime and accuracy vary depending on how the targeted class is selected. First, we compare the sequence- and sort-based method without early-stopping~($k=\abs{\Smy}$). Our proposed sort-based method yields a significant reduction in time without any loss in accuracy. Next, we consider the role of the hyper-parameter $k$. The sort-based method with early stopping successfully reduces the time significantly while compromising accuracy. Larger values of the hyper-parameter $k$ ensure higher success rates, but necessitate more time. However, there are cases where $k=5$ can significantly reduce the time with almost equal accuracy to that for $k=\abs{\Smy}$.

As a future policy in this field, we do not recommend the iterative method with a large $k$ for evaluation. Large $k$ is very time-consuming and requires high-performance computational resources for multiple evaluations, which would limit the number of researchers and hinder future development~\cite{GreenAI}. Instead, we recommend the iterative method with a small $k$, or the non-iterative method.

\subsection{Comparison of Non-Iterative Method}
\label{sec:ExpNon}
\cref{tab:non1,tab:non2} show the superclass accuracy extracted from max~(CW), sum~(CE), sum~(logit-CE), sum~(weighted-cw), and LSE~(CW) in 1 and 100 steps of PGD, respectively. Because the time taken by each non-iterative method does not vary significantly, it is omitted in \cref{tab:non1,tab:non2}. More detailed results of the non-iterative SAAs are listed in \cref{tab:ApNon}.

\cref{tab:non1,tab:non2} indicate that sum~(CE) is inefficient as an SAA. An adversarial attack by CE loss for a class $i\in\Smy$ decreases the class probability $p_i$ while increasing the $p_j$ of other classes $j\in\Y\backslash\{i\}$. The roles of increasing and decreasing the probability of each CE loss in sum~(CE) are in mutual conflict. Thus, sum~(CE) is not a suitable loss function for SAAs.

\cref{tab:non1,tab:non2} also shows that sum~(logit-CE) is not preferable for SAAs due to the gradient vanishing problem. When the probability of the targeted class is close enough to 1, the CE loss and its gradients approach 0, \ie, which causes gradient vanishing. Sum~(logit-CE) calculates the probability by summing all logits within the original superclass and applying the softmax function. When the superclass is composed of visual similarities, the logits $l_i~(i\in\Sy)$ tend to be positive, and their sum $\sum_{i\in\Sy}l_i$ tends to be greater than the original logit $l_y$. In this case, the probability in the sum~(logit-CE) $\frac{\e{\sum_{i\in\Sy}l_i}}{\sum_{I\in\SY}\e{\sum_{i\in I}l_i}}$ approaches 1 faster than the original probability $p_y$, \ie, which accelerate gradient vanishing. Therefore, sum~(logit-CE) is inefficient.

In a single-step, the LSE~(CW) is the best for SAAs. Max~(CW) does not work well with single-step because it reduces the only probability of the fine class $\pii{\Sy}$ and cannot reduce the superclass’s overall probability. In contrast, LSE- and sum-based methods incorporate the values of all fine classes into the loss function, and can decrease multiple probabilities, \ie, the superclass’s overall probability, by single-step. However, because sum~(weighted-CW) is somewhat less accurate than max~(CW), the incorporation of multiple fine class values is important.

For 100 steps, sum~(weighted-CW) method is the best method. Intuitively, sum~(weighted-CW) and LSE~(CW) may be expected to have similar robust accuracy in terms of summing multiple values of fine classes. However, the differences shown in \cref{tab:non1,tab:non2} indicate that SAA necessitates a careful selection of the loss function in accordance with the number of steps. One reason for the poor accuracy of max~(CW) in multi-steps is oscillation. When the maximum probability of one fine class $\pii{\Sy}$ decreases, the probability of another fine class $j\in\Sy\backslash\{\pii{\Sy}\}$ may increase. The sum- and LSE-based methods always calculate the gradients of all fine classes at each step, and can therefore achieve a higher success rate.

\tab{Comparison table of non-iterative SAAs in a single step. The best attack success rates are indicated in bold. Values represent superclass accuracy~(\%).}{tab:non1}{lrrrr}{
  \toprule
  method & \multicolumn{2}{r}{cifar-100} & \multicolumn{2}{r}{imagenet} \\ \cmidrule(lr){2-3} \cmidrule(lr){4-5}
  (1 step) & $\Mstd$ & $\Madv$ & $\Mstd$ & $\Madv$ \\ \midrule
  max~(cw) & 17.3\% & 51.8\% & 7.29\% & \tbf{47.1\%} \\
  sum~(ce) & 75.9\% & 73.2\% & 53.3\% & 53.1\% \\
  sum~(logit-ce) & 43.2\% & 62.4\% & 32.1\% & 58.7\% \\
  sum~(w-cw) & 24.6\% & 55.4\% & 14.8\% & 48.1\% \\
  lse~(cw) & \tbf{16.3\%} & \tbf{51.2\%} & \tbf{6.97\%} & \tbf{47.1\%} \\
  \bottomrule
}

\tab{Comparison table of non-iterative SAAs in 100 steps. The description is the same as in \cref{tab:non1}.}{tab:non2}{lrrrr}{
  \toprule
  method & \multicolumn{2}{r}{cifar-100} & \multicolumn{2}{r}{imagenet} \\ \cmidrule(lr){2-3} \cmidrule(lr){4-5}
  (100 steps.) & $\Mstd$ & $\Madv$ & $\Mstd$ & $\Madv$ \\ \midrule
  max~(cw) & \tbf{0.00\%} & 28.2\% & \tbf{0.00\%} & 24.5\% \\
  sum~(ce) & 77.2\% & 72.1\% & 50.3\% & 37.2\% \\
  sum~(logit-ce) & 38.7\% & 49.6\% & 19.8\% & 47.8\% \\
  sum~(w-cw) & \tbf{0.00\%} & \tbf{26.3\%} & \tbf{0.00\%} & \tbf{21.7\%} \\
  lse~(cw) & \tbf{0.00\%} & 27.8\% & \tbf{0.00\%} & 24.6\% \\
  \bottomrule
}

\section{Conclusion}
In this study, we considered superclass adversarial attacks~(SAAs) and superclass adversarial examples~(SAEs) as stepping stones to understanding higher-risk adversarial attacks. Our comprehensive analysis of SAAs~(an existing and 19 new methods) revealed iterative methods are strong but time-consuming, while non-iterative methods are fast but slightly weaker. The experimental results showed that among non-iterative methods, a variant of LSE- and sum-based methods achieves the highest attack success rate and the best efficiency in single- and multi-step attacks. In addition, we suggested that CE should not be incorporated into loss functions of non-iterative methods because of the conflicts between multiple CEs and the accelerated gradient vanishing problem. Our findings are not only important for the future development of SAAs, but can also be generalized to adversarial attacks with multiple targeted classes. In a future study, we further analyze what in a loss function of non-iterative SAA strongly affects a successful attack. Furthermore, we consider more threatening SAAs with a superclass configuration constructed by danger level rather than visual similarity.


\bibliography{bib/main,bib/AdvAtk,bib/AdvDef,bib/AdvMulti,bib/AdvPhy,bib/AdvThe,bib/Lib,bib/OOD}
\bibliographystyle{icml2022}

\newpage
\appendix
\onecolumn

\renewcommand{\thesection}{A\arabic{section}}
\renewcommand{\thetable}{A\arabic{table}}
\renewcommand{\thefigure}{A\arabic{figure}}

\setcounter{section}{0}
\setcounter{table}{0}
\setcounter{figure}{0}

\section{Additional Related Work}
\label{sec:ApRelated}
\cref{sec:related} describes studies that relate to standard adversarial attacks and SAAs. In this section, top-k and multi-label classification attacks are discussed as adversarial attacks that involve multiple classes.

Multi-label classification involves assigning multiple classes to a single instance~(\eg, an image)~\cite{song2018multi,zhou2020generating,ghamizi2021evasion,melacci2021domain,yang2021characterizing,yang2021attack,zhou2021hiding}. Some studies have proposed adversarial attacks in multi-label classification~\cite{song2018multi,zhou2020generating} as an extension of the C\&W attack or DeepFool~\cite{DeepFool}. Song~\etal~leveraged the relationships between labels for multi-label classification attacks~\cite{song2018multi} motivated by multi-label ranking loss~\cite{schapire2000boostexter}. Some studies have demonstrated empirical research of adversarial examples that hide all labels~\cite{zhou2021hiding}, theoretical attackability~\cite{yang2021characterizing,yang2021attack}, adversarial defense~\cite{melacci2021domain}, and relationships with steganography~\cite{ghamizi2021evasion} in the multi-label classification attacks. This study considers single-label classification wherein one class is assigned to each instance. In addition, information concerning the relationships between the classes, which may be useful for attacks, is not used.

Some studies have considered adversarial attacks focusing on the top-k prediction of the classifier~\cite{tursynbek2022geometry,zhang2020learning,hu2021tkml}. Tursynbek~\etal~considered an attack in which the true label of an image was excluded from the top-k prediction of the classifier~\cite{tursynbek2022geometry}. Zhang~\etal~proposed an attack that replaces the top-k of the classifier with the class specified by an adversary~\cite{zhang2020learning}. Hu~\etal~considered an adversarial attack on top-k multi-label learning, where an instance is associated with a non-empty subset of labels, and the output of the system is the top-k predicted labels~\cite{hu2021tkml}. This study does not focus on top-k predictions, with the goal being to prevent top-1 predictions from being included in the original superclass.

Although some multi-classification and top-k attacks can be applied to SAA with some modifications~\cite{song2018multi,zhou2020generating,tursynbek2022geometry,zhang2020learning,hu2021tkml}, they were not considered in this study for the following reasons. One of the goals, here, is to compare several loss functions under identical conditions to determine the configuration best suited for SAAs. Therefore, these studies deviated from our experimental settings. Furthermore, one of the nine proposed methods, the sum~(CW) method~(cf. \cref{sec:MethodSingle}), is essentially based on the same idea~(\ie, idea based on C\&W attack) as~\cite{song2018multi,zhang2020learning,hu2021tkml}, and can therefore, be used as a substitute. \cite{song2018multi,zhou2020generating,tursynbek2022geometry} proposed geometric methods, which may be explored in future work.

\section{Experimental Settings}
\label{sec:ApSettings}
This section describes the experimental settings in detail.

\paragraph{Datasets.}
The superclass settings for CIFAR-100 and ImageNet followed~\cite{CIFAR10} and~\cite{ImageNetSuperclass}, respectively\footnote{The configuration details of ImageNet is provided at \url{https://github.com/s-kumano/imagenet-superclass}.}. The number of superclasses given in CIFAR-100 was 20, and each superclass included five~(fine) classes. The number of superclasses provided in ImageNet was 558, the average number of elements 1.79, and the standard deviation of the elements 1.57.

\paragraph{Models.}
For CIFAR-100, we trained WideResNet-40-10~\cite{WideResNet}. For ImageNet, we used pretrained WideResNet-50-2~(normally trained model from~\cite{pytorch} and adversarially trained one from~\cite{salman2020adversarially}).

\paragraph{CIFAR-100 Training.}
The number of epochs was 200 and the batch size was 128. The optimizer was a stochastic gradient descent with a learning rate of 0.1, momentum of 0.9, and weight decay of 2e-4, and the learning rate was multiplied by 0.1 at epochs 100 and 150, respectively. In adversarial training, a PGD with 10 steps was used: $l_\infty$ norm, 8/255 constraint of distance $\epsilon$, and 2/255 step size $\alpha$. As in~\cite{FriendAT}, updates of adversarial examples were stopped when an attack was successful to reduce the training time. Note that adversarial training was conducted using standard adversarial examples for fine class misclassification.

\paragraph{Validation.}
An NVIDIA A100-PCIE-40GB was used for all evaluations. The batch sizes were 2,048 for CIFAR-100 and 512 for ImageNet. In ImageNet, 5,000 images were randomly extracted. The based PGD settings are 100 steps: $l_\infty$ norm, 8/255 distance constraint $\epsilon$, 2/255 step size $\alpha$, with random initialization by a uniform distribution $U(-\epsilon,\epsilon)$. In the iterative method, random initialization was performed for each targeted attack. The updates of adversarial examples were stopped when the attack was successful, that is, when superclass misclassification was achieved.

\section{Experimental Results}
\cref{tab:ApStd} shows the robust accuracy under the standard adversarial attacks with four loss functions. 
Robust accuracy varies across loss functions, but attacks with them cause fewer superclass misclassifications than SAAs. These results indicate that standard adversarial attacks cause misclassification between similar fine classes and are not at high risk for DNN-based systems.

\cref{tab:ApIterative} shows the results of iterative SAAs. The larger the $k$, the higher is the attack success rate; however, the operation will take a longer time. Note that the results on ImageNet are only a sample of 5,000, and iterative SAAs with large $k$ for all images in ImageNet will take a considerable amount of time. Owing to different random initialization, the superclass accuracy under sort~(weighted-CW) with $k=5$ was larger than that with $k=4$.

\cref{tab:ApNon} shows the results of non-iterative SAAs with one, 10, and 100 steps. In a single step, LSE~(CW) achieves a high attack success rate. This is because the gradients of all fine classes in the same superclass are calculated in a single step, and are weighted more efficiently than the other sum- and LSE-based methods. In 100 steps, the sum-based methods were stronger than the other two methods. In particular, the sum~(weighted-CW) in 100 steps achieved a high attack success rate in a short time. The sum~(CE) and LSE~(CE) in 10 steps were weaker than those in a single step. This is attributed to the difference in step size.

\tab{Comparison table of the standard adversarial attacks. The best attack success rates are indicated in bold. Values represent accuracy~(\%) and time~(minutes).}{tab:ApStd}{lrrrr}{
  \toprule
  \multirow{2}{*}{loss} & \multicolumn{2}{r}{cifar-100} & \multicolumn{2}{r}{imagenet} \\ \cmidrule(lr){2-3} \cmidrule(lr){4-5}
  & $\Mstd$ & $\Madv$ & $\Mstd$ & $\Madv$ \\ \midrule
  ce & 30.1\%~(1.00min) & \tbf{43.1\%~(2.00min)} & 14.4\%~(1.00min) & \tbf{26.1\%~(2.00min)} \\
  cw & 34.7\%~(2.00min) & 47.6\%~(2.00min) & 19.1\%~(1.00min) & 32.8\%~(2.00min) \\
  prob-cw & 34.3\%~(1.00min) & 46.2\%~(2.00min) & 20.7\%~(1.00min) & 29.8\%~(2.00min) \\
  weighted-cw & \tbf{26.5\%~(1.00min)} & 44.8\%~(2.00min) & \tbf{12.2\%~(1.00min)} & 29.6\%~(2.00min) \\
  \bottomrule
}

\tab{Comparison table of the iterative SAAs. The best attack success rates are indicated in bold. Values represent accuracy~(\%) and time~(minutes).}{tab:ApIterative}{lllrrrr}{
  \toprule
  \multirow{2}{*}{$k$} & \multirow{2}{*}{method} & \multirow{2}{*}{loss} & \multicolumn{2}{r}{cifar-100} & \multicolumn{2}{r}{imagenet} \\ \cmidrule(lr){4-5} \cmidrule(lr){6-7}
  & & & $\Mstd$ & $\Madv$ & $\Mstd$ & $\Madv$ \\ \midrule
  1 & sort & ce & \tbf{0.00\%~(0.00min)} & \tbf{28.2\%~(1.00min)} & \tbf{0.00\%~(0.00min)} & 27.5\%~(2.00min) \\
  & & cw & \tbf{0.00\%~(0.00min)} & 28.4\%~(1.00min) & \tbf{0.00\%~(0.00min)} & 25.1\%~(2.00min) \\
  & & prob-cw & 5.13\%~(0.00min) & 32.6\%~(2.00min) & 3.26\%~(0.00min) & \tbf{23.1\%~(2.00min)} \\
  & & w-cw & 0.06\%~(0.00min) & 30.6\%~(2.00min) & \tbf{0.00\%~(0.00min)} & 23.2\%~(2.00min) \\ \midrule
  2 & sort & ce & \tbf{0.00\%~(0.00min)} & \tbf{26.5\%~(3.00min)} & \tbf{0.00\%~(0.00min)} & 24.0\%~(4.00min) \\
  & & cw & \tbf{0.00\%~(0.00min)} & 26.8\%~(3.00min) & \tbf{0.00\%~(0.00min)} & 22.8\%~(4.00min) \\
  & & prob-cw & 4.10\%~(0.00min) & 32.1\%~(3.00min) & 2.97\%~(1.00min) & \tbf{21.9\%~(3.00min)} \\
  & & w-cw & 0.06\%~(0.00min) & 30.2\%~(3.00min) & \tbf{0.00\%~(0.00min)} & 22.1\%~(4.00min) \\ \midrule
  3 & sort & ce & \tbf{0.00\%~(0.00min)} & \tbf{26.1\%~(4.00min)} & \tbf{0.00\%~(0.00min)} & 22.5\%~(6.00min) \\
  & & cw & \tbf{0.00\%~(0.00min)} & 26.3\%~(4.00min) & \tbf{0.00\%~(0.00min)} & 21.8\%~(5.00min) \\
  & & prob-cw & 3.68\%~(1.00min) & 32.0\%~(5.00min) & 2.77\%~(2.00min) & \tbf{21.5\%~(5.00min)} \\
  & & w-cw & 0.06\%~(0.00min) & 30.0\%~(5.00min) & \tbf{0.00\%~(0.00min)} & 21.7\%~(5.00min) \\ \midrule
  4 & sort & ce & \tbf{0.00\%~(0.00min)} & \tbf{25.8\%~(6.00min)} & \tbf{0.00\%~(0.00min)} & 22.2\%~(7.00min) \\
  & & cw & \tbf{0.00\%~(0.00min)} & 26.1\%~(6.00min) & \tbf{0.00\%~(0.00min)} & 21.4\%~(7.00min) \\
  & & prob-cw & 3.39\%~(1.00min) & 32.0\%~(7.00min) & 2.74\%~(2.00min) & \tbf{21.3\%~(7.00min)} \\
  & & w-cw & 0.03\%~(1.00min) & 29.9\%~(7.00min) & \tbf{0.00\%~(0.00min)} & 21.4\%~(7.00min) \\ \midrule
  5 & sort & ce & \tbf{0.00\%~(0.00min)} & \tbf{25.7\%~(7.00min)} & \tbf{0.00\%~(0.00min)} & 21.9\%~(9.00min) \\
  & & cw & \tbf{0.00\%~(0.00min)} & 25.9\%~(7.00min) & \tbf{0.00\%~(0.00min)} & \tbf{21.2\%~(8.00min)} \\
  & & prob-cw & 3.31\%~(1.00min) & 31.9\%~(9.00min) & 2.67\%~(3.00min) & \tbf{21.2\%~(8.00min)} \\
  & & w-cw & 0.05\%~(1.00min) & 29.8\%~(8.00min) & \tbf{0.00\%~(0.00min)} & 21.4\%~(8.00min) \\ \midrule
  $\abs{\Sy}$ & seq. & ce & \tbf{0.00\%~(0.00min)} & 25.4\%~(171min) & \tbf{0.00\%~(0.00min)} & 20.6\%~($203\times10^1$min) \\
  & sort & ce & \tbf{0.00\%~(0.00min)} & \tbf{25.4\%~(138min)} & \tbf{0.00\%~(0.00min)} & \tbf{20.6\%~($\bf{153\times10^1}$min)} \\
  \bottomrule
}

\tab{Comparison table of the iterative SAAs. Values represent accuracy~(\%) and time~(minutes). The best attack success rates in each step are indicated in bolded and they in each step and broadly categorized method~(max, sum, and LSE) are in underscore.}{tab:ApNon}{lllrrrr}{
  \toprule
  \multirow{2}{*}{steps} & \multirow{2}{*}{method} & \multirow{2}{*}{loss} & \multicolumn{2}{r}{cifar-100} & \multicolumn{2}{r}{imagenet} \\ \cmidrule(lr){4-5} \cmidrule(lr){6-7}
  & & & $\Mstd$ & $\Madv$ & $\Mstd$ & $\Madv$ \\ \midrule
  1 & max & ce & 32.1\%~(0.00min) & 57.0\%~(0.00min) & 21.9\%~(0.00min) & 49.9\%~(0.00min) \\
  & & cw & \ul{17.3\%~(0.00min)} & \ul{51.8\%~(0.00min)} & \ul{7.29\%~(0.00min)} & \ulbf{47.1\%~(0.00min)} \\
  & & prob-cw & 30.3\%~(0.00min) & 55.2\%~(0.00min) & 20.2\%~(0.00min) & 48.0\%~(0.00min) \\
  & & weighted-cw & 27.8\%~(0.00min) & 57.3\%~(0.00min) & 17.7\%~(0.00min) & 48.3\%~(0.00min) \\ \cmidrule(lr){2-7}
  & sum & ce & 75.9\%~(0.00min) & 73.2\%~(0.00min) & 53.3\%~(0.00min) & 53.1\%~(0.00min) \\
  & & logit-ce & 43.2\%~(0.00min) & 62.4\%~(0.00min) & 32.1\%~(0.00min) & 58.7\%~(0.00min) \\
  & & prob-ce & 21.6\%~(0.00min) & 52.6\%~(0.00min) & 12.5\%~(0.00min) & 49.0\%~(0.00min) \\
  & & cw & \ul{19.5\%~(0.00min)} & 57.4\%~(0.00min) & 12.8\%~(0.00min) & 50.6\%~(0.00min) \\
  & & prob-cw & 20.5\%~(0.00min) & \ul{52.0\%~(0.00min)} & \ul{11.8\%~(0.00min)} & \ul{47.8\%~(0.00min)} \\
  & & weighted-cw & 24.6\%~(0.00min) & 55.4\%~(0.00min) & 14.8\%~(0.00min) & 48.1\%~(0.00min) \\ \cmidrule(lr){2-7}
  & lse & ce & 77.8\%~(0.00min) & 75.2\%~(0.00min) & 55.9\%~(0.00min) & 56.0\%~(0.00min) \\
  & & prob-ce & 32.3\%~(0.00min) & 55.0\%~(0.00min) & 20.9\%~(0.00min) & 49.1\%~(0.00min) \\
  & & cw & \ulbf{16.3\%~(0.00min)} & \ulbf{51.2\%~(0.00min)} & \ulbf{6.97\%~(0.00min)} & \ulbf{47.1\%~(0.00min)} \\
  & & prob-cw & 27.6\%~(0.00min) & 52.9\%~(0.00min) & 17.6\%~(0.00min) & 47.7\%~(0.00min) \\
  & & weighted-cw & 27.5\%~(0.00min) & 57.0\%~(0.00min) & 17.6\%~(0.00min) & 48.1\%~(0.00min) \\ \midrule
  10 & max & ce & 7.72\%~(0.00min) & 40.7\%~(0.00min) & 3.53\%~(0.00min) & 31.9\%~(0.00min) \\
  & & cw & \ulbf{0.10\%~(0.00min)} & \ul{33.5\%~(0.00min)} & \ulbf{0.00\%~(0.00min)} & 29.9\%~(0.00min) \\
  & & prob-cw & 6.58\%~(0.00min) & 38.2\%~(0.00min) & 3.28\%~(0.00min) & \ul{28.7\%~(0.00min)} \\
  & & weighted-cw & 1.15\%~(0.00min) & 38.3\%~(0.00min) & \ulbf{0.00\%~(0.00min)} & 28.8\%~(0.00min) \\ \cmidrule(lr){2-7}
  & sum & ce & 77.2\%~(0.00min) & 72.1\%~(0.00min) & 50.5\%~(0.00min) & 41.5\%~(0.00min) \\
  & & logit-ce & 38.7\%~(0.00min) & 52.3\%~(0.00min) & 20.0\%~(0.00min) & 51.4\%~(0.00min) \\
  & & prob-ce & 6.84\%~(0.00min) & 33.1\%~(0.00min) & 4.52\%~(0.00min) & 31.0\%~(0.00min) \\
  & & cw & \ul{0.20\%~(0.00min)} & 40.9\%~(0.00min) & \ulbf{0.00\%~(0.00min)} & 35.0\%~(0.00min) \\
  & & prob-cw & 5.90\%~(0.00min) & \ulbf{32.7\%~(0.00min)} & 3.95\%~(0.00min) & \ulbf{28.4\%~(0.00min)} \\
  & & weighted-cw & 4.00\%~(0.00min) & 34.6\%~(0.00min) & \ulbf{0.00\%~(0.00min)} & 28.8\%~(0.00min) \\ \cmidrule(lr){2-7}
  & lse & ce & 79.4\%~(0.00min) & 74.5\%~(0.00min) & 53.2\%~(0.00min) & 47.2\%~(0.00min) \\
  & & prob-ce & 11.8\%~(0.00min) & 36.8\%~(0.00min) & 4.72\%~(0.00min) & 31.0\%~(0.00min) \\
  & & cw & \ulbf{0.10\%~(0.00min)} & \ul{33.1\%~(0.00min)} & \ulbf{0.00\%~(0.00min)} & 30.2\%~(0.00min) \\
  & & prob-cw & 7.15\%~(0.00min) & 34.6\%~(0.00min) & 3.70\%~(0.00min) & 29.1\%~(0.00min) \\
  & & weighted-cw & 0.83\%~(0.00min) & 37.3\%~(0.00min) & \ulbf{0.00\%~(0.00min)} & \ul{28.7\%~(0.00min)} \\ \midrule
  100 & max & ce & 5.70\%~(0.00min) & 34.6\%~(2.00min) & 3.51\%~(0.00min) & 24.8\%~(2.00min) \\
  & & cw & \ulbf{0.00\%~(0.00min)} & \ul{28.2\%~(1.00min)} & \ulbf{0.00\%~(0.00min)} & 24.5\%~(2.00min) \\
  & & prob-cw & 4.98\%~(0.00min) & 32.3\%~(1.00min) & 3.26\%~(0.00min) & 22.2\%~(2.00min) \\
  & & weighted-cw & 0.03\%~(0.00min) & 30.0\%~(1.00min) & \ulbf{0.00\%~(0.00min)} & \ul{21.9\%~(2.00min)} \\ \cmidrule(lr){2-7}
  & sum & ce & 77.2\%~(4.00min) & 72.1\%~(3.00min) & 50.3\%~(3.00min) & 37.2\%~(3.00min) \\
  & & logit-ce & 38.7\%~(2.00min) & 49.6\%~(2.00min) & 19.8\%~(1.00min) & 47.8\%~(3.00min) \\
  & & prob-ce & 6.61\%~(0.00min) & 27.1\%~(1.00min) & 4.52\%~(0.00min) & 23.7\%~(2.00min) \\
  & & cw & \ulbf{0.00\%~(0.00min)} & 35.4\%~(2.00min) & \ulbf{0.00\%~(0.00min)} & 29.0\%~(2.00min) \\
  & & prob-cw & 5.75\%~(0.00min) & 27.1\%~(1.00min) & 3.95\%~(0.00min) & 21.8\%~(2.00min) \\
  & & weighted-cw & \ulbf{0.00\%~(0.00min)} & \ulbf{26.3\%~(1.00min)} & \ulbf{0.00\%~(0.00min)} & \ulbf{21.7\%~(2.00min)} \\ \cmidrule(lr){2-7}
  & lse & ce & 79.4\%~(4.00min) & 74.5\%~(4.00min) & 53.0\%~(3.00min) & 43.2\%~(3.00min) \\
  & & prob-ce & 10.6\%~(0.00min) & 30.2\%~(1.00min) & 4.67\%~(0.00min) & 23.7\%~(2.00min) \\
  & & cw & \ulbf{0.00\%~(0.00min)} & \ul{27.8\%~(1.00min)} & \ulbf{0.00\%~(0.00min)} & 24.6\%~(2.00min) \\
  & & prob-cw & 6.36\%~(0.00min) & 28.5\%~(1.00min) & 3.68\%~(0.00min) & 22.9\%~(2.00min) \\
  & & weighted-cw & 0.01\%~(0.00min) & 28.7\%~(1.00min) & \ulbf{0.00\%~(0.00min)} & \ul{21.8\%~(2.00min)} \\
  \bottomrule
}

\end{document}